# Polish - English Speech Statistical Machine Translation Systems for the IWSLT 2013.

*Krzysztof Wołk, Krzysztof Marasek*

Multimedia Department
Polish Japanese Institute of Information Technology, Koszykowa 86, 02-008 Warsaw
kwolk@pjwstk.edu.pl, kmarasek@pjwstk.edu.pl

## Abstract

This research explores the effects of various training settings from Polish to English Statistical Machine Translation system for spoken language. Various elements of the TED parallel text corpora for the IWSLT 2013 evaluation campaign were used as the basis for training of language models, and for development, tuning and testing of the translation system. The BLEU, NIST, METEOR and TER metrics were used to evaluate the effects of data preparations on translation results. Our experiments included systems, which use stems and morphological information on Polish words. We also conducted a deep analysis of provided Polish data as preparatory work for the automatic data correction and cleaning phase.

## 1. Introduction

Polish is one of the most complex West-Slavic languages, which represents a serious challenge to any SMT system. The grammar of the Polish language, with its complicated rules and elements, together with a big vocabulary (due to complex declension) are the main reasons for its complexity. Furthermore, Polish has 7 cases and 15 gender forms for nouns and adjectives, with additional dimensions for other word classes.

This greatly affects the data and data structure required for statistical models of translation. The lack of available and appropriate resources required for data input to SMT systems presents another problem. SMT systems should work best in specified, not too wide text domains and will not perform well for general use. Good quality parallel data, especially in a required domain has low availability. In general, Polish and English differ also in syntax. English is a positional language, which means that the syntactic order (the order of words in a sentence) plays a very important role, particularly due to limited inflection of words (e.g. lack of declension endings). Sometimes, the position of a word in a sentence is the only indicator of the sentence meaning. In the English sentence, the subject group comes before the predicate, so the sentence is ordered according to the Subject-Verb-Object (SVO) schema. In Polish, however, there is no specific word order imposed and the word order has no decisive influence on the understanding of the sentence. One can express the same thought in several ways, which is not possible in English. For example, the sentence „I bought myself a new car." can be written in Polish as „Kupiłem sobie nowy samochód.", or "Nowy samochód sobie kupiłem.", or "Sobie kupiłem nowy samochód.", or „Samochód nowy sobie kupiłem." Differences in potential sentence orders make the translation process more complex, especially when working on a phrase-model with no additional lexical information.

As a result the progress in the development of SMT systems for Polish is substantially slower as compared to other languages. The aim of this work is to create an SMT system for translation from Polish to (and the other way round, i.e. from English to Polish) to address the IWSLT 2013 [2] evaluation campaign requirements. This paper is structured as follows: Section 2 explains the Polish data preparation. Section 3 presents the English language issues. Section 4 describes the translation evaluation methods. Section 5 discusses the results. Sections 6 and 7 summarize potential implications and future work.

## 2. Preparation of the Polish data

The Polish data in the TED talks (about 15 MB) include almost 2 million words that are not tokenized. The transcripts themselves are provided as pure text encoded with UTF-8 and the transcripts are prepared by the IWSLT team [3]. In addition, they are separated into sentences (one per line) and aligned in language pairs.

It should be emphasized that both automatic and manual preprocessing of this training information was required. The extraction of the transcription data from the provided XML files ensured an equal number of lines for English and Polish. However, some of the discrepancies in the text parallelism could not be avoided. These discrepancies are mainly repetitions of the Polish text not included in the English text.

Another problem is that TED 2013 data is full of errors. Let us first take spelling errors that artificially increase the dictionary size and make the statistics worse. We took a very large Polish dictionary [23] that consists of 2,532,904 different words. Then, we created a dictionary from TED 2013 data and it consisted of 92,135 unique words. Intersection of those 2 dictionaries resulted in a new dictionary containing 58,393 words. It means that in TED 2013 we found 33742 words that do not exist in Polish (spelling errors or named entities). This is as much as 36.6% of the whole TED Polish vocabulary.

To verify that, we conducted a manual analysis on a sample of the first 300 lines from the TED corpora. We found that there were 4268 words containing a total of 35 kinds of spelling errors that occurred many times. But what we found to be more problematic was that there were sentences with odd nesting, such as:

Part A, Part A, Part B, Part B.

e.g.

"Ale będę starał się udowodnić, że mimo złożoności, Ale będę starał się udowodnić, że mimo złożoności, istnieją pewne rzeczy pomagające w zrozumieniu. istnieją pewne rzeczy pomagające w zrozumieniu."

We can see that some parts (words or full phrases or even whole sentences) were duplicated. Furthermore, there are segments containing repeated whole sentences inside one segment. For instance:

Sentence A. Sentence A.
e.g.

"Zakumulują się u tych najbardziej pijanych i skąpych. Zakumulują się u tych najbardziej pijanych i skąpych."

or:

Part A, Part B, Part B, Part C
e.g.

" Matka może się ponownie rozmnażać, ale jak wysoką cenę płaci, przez akumulację toksyn w swoim organizmie - przez akumulację toksyn w swoim organizmie - śmierć pierwszego młodego."

We identified 51 out of 300 segments that were mistaken in such way. Overall, in the sample test set we found that we got about 10% of spelling errors and about 17% of insertion errors. However, it must be noted that we simply took the first 300 lines, but in the whole text there are places where more problems occur. So, to some extent, this confirms that there are problems related to the dictionary.

Additionally, there are a number of English names, words and phrases (nor translated) present in the Polish text. There are also some sentences originating from different languages (e.g., German and French). Additionaly some translations are just incorrect or too indirect with not enough precision in translation, e.g. "And I remember there sitting at my desk thinking, Well, I know this. This is a great scientific discovery. " was translated into " Pamiętam, jak pomyślałem: To wyjątkowe, naukowe odkrycie." And the correct translation would be "Pamiętam jak siedząc przy biurku pomyślałem, dobrze, wiem to. To jest wielkie naukowe odkrycie".

The size of the vocabulary is 92,135 Polish unique words and 41,684 English unique words. The disproportionate vocabulary sizes are also a challenge especially in translation from English to Polish.

Another serious problem (especially for Statistical Machine Translation) that we found was that English sentences were translated in an improper manner.

There were four main problems:

1. Repetitions – when part of the text is repeated several times after translation, i.e.
   a. EN: Sentence A. Sentence B.
   b. PL: Translated Sentence A. Translated Sentence B. Translated Sentence B. Translated Sentence B.
2. Wrong usage of words – when one or more words used for the Polish translation change slightly the meaning of the original English sentence, i.e.
   a. EN: We had these data a few years ago.
   b. PL (the proper meaning of the polish sentence): We've been delivered these data a few years ago.

3. Indirect translations, usage of metaphors – when the Polish translation uses different wording in order to preserve the meaning of the original sentence, especially when the exact translation would result in a sentence with no sense. Many metaphors are translated this way.
4. Translations that are not precise enough – when the translated fragment does not contain all the details of the original sentence, but only its overall meaning is the same.

Looking at the style of the translated text, it can be concluded that the text was translated by several people who translated it independently of one another. The text was divided between them and then the fragments of it were merged into one, thus:

- In some places, the text looks as if it was not translated by a human, but by an automatic system instead.

- Some paragraphs contain a lot of metaphors, which will certainly interfere with the subsequent translation. The translations are not direct.

- We also found some problems with encoding of Polish characters and with usage some strange symbols in the text ex. Ҳ, Ψ, ∑, ℧, ∏, , etc.

Another problem is that the TED Talks do not have any specific domain. Statistical Machine Translation by definition works best when very specific domain data is used. The data we have is a mix of various, unrelated topics. This is most likely the reason why we cannot expect big improvements with this data.

There is not much focus on Polish in the campaign, so there is almost no data in Polish in comparison to a huge amount of data in, for example, French or German. What is more, provided Polish samples are not only small, but also in a different domain, which does not enrich a required language model well enough. At first we used perplexity measurement metrics to determine the data we got. Some of it we were able to obtain from the project page, some from another project and the rest was collected manually using web crawlers. We created those corpora and used them according to the permission from organizers [22]. What we created was:

- A Polish – English dictionary (bilingual parallel)

- Additional (newer) TED Talks data sets not included in the original train data (we crawled bilingual data and created a corpora from it) (bilingual parallel)

- E-books (monolingual PL + monolingual EN)

- Euro News Data (bilingual parallel)

- Proceedings of UK Lords (monolingual EN)

- Subtitles for movies and TV series (monolingual PL)

- Parliament and senate proceedings (monolingual PL)

"Other" in the table below stands for many very small models merged together. We show here the perplexity values and the perplexity values with no smoothing (PPL in Table I) of those language models smoothed with the Kneser-Ney algorithm (PPL+KN in Table I). We used the MITLM toolkit for that evaluation. As an evaluation set we used dev2010 data, which was used for tuning. Its dictionary covers 2861 different words.

*Table 1*: Data Perplexities for dev2010 data set

| Data set | Dictionary | PPL | PPL + KN |
|---|---|---|---|
| Baseline train.en | 44,052 | 221 | 223 |
| EMEA | 30,204 | 1738 | 1848 |
| KDE4 | 34,442 | 890 | 919 |
| ECB | 17,121 | 837 | 889 |
| OpenSubtitles | 343,468 | 388 | 415 |
| EBOOKS | 528,712 | 405 | 417 |
| EUNEWS | 34,813 | 430 | 435 |
| NEWS COMM | 62,937 | 418 | 465 |
| EUBOOKSHOP | 167,811 | 921 | 950 |
| UN TEXTS | 175,007 | 681 | 714 |
| UK LORDS | 215,106 | 621 | 644 |
| NEWS 2010 | 279,039 | 356 | 377 |
| GIGAWORD | 287,096 | 582 | 610 |
| DICTIONARY | 39,214 | 8629 | 8824 |
| OTHER | 13,576 | 492 | 499 |
| TEDDL | 47,015 | 277 | 277 |

EMEA are texts from the European Medicines Agency, KDE4 is a localization file of that user GUI, ECB stands for European Central Bank corpus, OpenSubtitles are movies and TV series subtitles, EUNEWS is a web crawl of the euronews.com web page and EUBOOKSHOP comes from bookshop.europa.eu. Lastly bilingual TEDDL is additional TED data. As can be seen from the table above, all additional data is much worse than the files provided in the baseline system, so no major improvements based only on data could be anticipated.

Before the use of a training translation model, preprocessing that included removal of long sentences (set to 80 words) had to be performed. The Moses toolkit scripts[6] were used for this purpose. Moses is an open-source toolkit for statistical machine translation which supports linguistically motivated factors, confusion network decoding, and efficient data formats for translation models and language models. In addition to the SMT decoder, the toolkit also includes a wide variety of tools for training, tuning and applying the system to many translation tasks. In addition, the text in the TED data set had to be repaired in a number of ways to correct spelling errors and grammar errors, ensure that there was only one sentence on each line, remove language translations that were not of interest, remove HTML and XML tags within text, remove of strange symbols not existing in a specific language and repetitions of words and sentences.

The final processing included 134,678 lines from the Polish to English corpus. However, the disproportionate vocabulary sizes remained, with 41,163 English words and 92,135 Polish words. One of the solutions to this problem (according to work of Bojar [7]) was to use stems instead of surface forms that reduced the Polish vocabulary size to 40,346. Such a solution also requires a creation of an SMT system from Polish stems to plain Polish. Subsequently, morphosyntactic tagging, using the Wroclaw Natural Language Processing (NLP) tools (nlp.pwr.wroc.pl), was included as an additional information source for the SMT system preparation. It can be also used as a first step for implementing a factored SMT system that, unlike a phrase-based system, includes morphological analysis, translation of lemmas and features as well as generation of surface forms. Incorporating additional linguistic information should effectively improve translation performance [8].

## 2.1. Polish stem extraction

As previously mentioned, stems extracted from Polish words are used instead of surface forms to overcome the problem of the huge difference in vocabulary sizes. Keeping in mind that in half of the experiments the target language was English in the form of normal sentences, it was not necessary to introduce models for converting the stems to the appropriate grammatical forms; however it will be part of our future work in translation into Polish. For Polish stem extraction, a set of natural language processing tools available at http://nlp.pwr.wroc.pl was used [9]. These tools can be used for:

- Tokenization
- Morphosyntactic analysis
- Shallow parsing as chunking
- Text transformation into the featured vectors

The following two components are also used:

- MACA –a universal framework used to connect the different morphological data
- WCRFT – this framework combines conditional random fields and tiered tagging

These tools used in sequence provide an XML output. It includes a surface form of the tokens, stems and morphosyntactic tags. An example of such data is given in section 2.2.

## 2.2. Morphosyntactic element tagging with standard tools

Wroclaw's tools were used to tag morphosyntactic elements. More precise tagging can be achieved with these settings. In addition, every tag in this tagset consists of specific grammatical classes with specific values for particular attributes. Furthermore, these grammatical classes include attributes with values that require additional specification. For example, nouns require numbers while adverbs require an appropriate degree of an attribute. This causes segmentation of the input data, including tokenization of words in a different way as compared to the Moses tools. On the other hand, this causes problems with building parallel corpora. This can be solved by placing markers at the end of input lines.

In the following example, where pl.gen. "men" is derived from sin.nom."człowiek" (*man*) or pl.nom. "ludzie" (*people*), it can be demonstrated how one tag is used where, in the most difficult cases, more possible tags are provided.

```
<tok>
<orth>ludzi</orth>
<lex disamb="1"><base>człowiek</base>
<ctag>subst:pl:gen:m1</ctag></lex>
<lex disamb="1"> <base>ludzie</base>
<ctag>subst:pl:gen:m1</ctag></lex>
</tok>
```

In this example, only one form (the first stem) is used for further processing.

We developed an XML extractor tool to generate three different corpora for the Polish language data:

- Words in the infinitive form
- Subject-Verb-Object (SVO) word order
- both the infinitive form and the SVO word order

This allows experiments with those preprocessing techniques.

Moreover, some of the NLP tools use the Windows-1250 Eastern Europe Character Encoding, which requires a conversion of information to and from the UTF-8 encoding that is commonly used in other, standard tools.

## 3. English Data Preparation

The preparation of the English data was definitively less complicated than for Polish. We developed a tool to clean the English data by removing foreign words, strange symbols, etc. Compare to polish english data contained significantly less errors. Nonetheless some problems needed to be removed, most problematic were translations into languages other than english, strange UTF-8 symbols. We also found few duplications and insertions inside single segment.

## 4. Evaluation Methods

Metrics are necessary to measure the quality of translations produced by the SMT systems. For this, various automated metrics are available to compare SMT translations to high quality human translations. Since each human translator produces a translation with different word choices and orders, the best metrics measure SMT output against multiple reference human translations. Among the commonly used SMT metrics are: Bilingual Evaluation Understudy (BLEU), the U.S. National Institute of Standards & Technology (NIST) metric, the Metric for Evaluation of Translation with Explicit Ordering (METEOR), and Translation Error Rate (TER). These metrics will now be briefly discussed. [10]

BLEU was one of the first metrics to demonstrate high correlation with reference human translations. The general approach for BLEU, as described in [9], is to attempt to match variable length phrases to reference translations. Weighted averages of the matches are then used to calculate the metric. The use of different weighting schemes leads to a family of BLEU metrics, such as the standard BLEU, Multi-BLEU, and BLEU-C. [11]

As discussed in [11], the basic BLEU metric is:

$$BLEU = P_B \exp\left(\sum_{n=0}^{N} w_n \log p_n\right)$$

where $p_n$ is an $n$-gram precision using $n$-grams up to length N and positive weights $w_n$ that sum to one. The brevity penalty $P_B$ is calculated as:

$$P_B = \begin{cases} 1, & c > r \\ e^{(1-r/c)}, & c \leq r \end{cases}$$

where $c$ is the length of a candidate translation, and $r$ is the effective reference corpus length. [9]

The standard BLEU metric calculates the matches between $n$-grams of the SMT and human translations, without considering position of the words or phrases within the texts. In addition, the total count of each candidate SMT word is limited by the corresponding word count in each human reference translation. This avoids bias that would enable SMT systems to overuse high confidence words in order to boost their score. BLEU applies this approach to texts sentence by sentence, and then computes a score for the overall SMT output text. In doing this, the geometric mean of the individual scores is used, along with a penalty for excessive brevity in translation. [9]

The NIST metric seeks to improve the BLEU metric by valuing information content in several ways. It takes the arithmetic versus geometric mean of the $n$-gram matches to reward good translation of rare words. The NIST metric also gives heavier weights to rarer words. Lastly, it reduces the brevity penalty when there is a smaller variation in translation length. This metric has demonstrated that these changes improve the baseline BLEU metric. [12]

The METEOR metric, developed by the Language Technologies Institute of Carnegie Mellon University, is also intended to improve the BLEU metric. METEOR rewards recall by modifying the BLEU brevity penalty, takes into account higher order $n$-grams to reward matches in word order, and uses arithmetic vice geometric averaging. For multiple reference translations, METEOR reports the best score for word-to-word matches. Banerjee and Lavie [13] describe this metric in detail.

As found in [13], this metric is calculated as follows:

$$METEOR = \left(\frac{10\,P\,R}{R + 9\,P}\right)(1 - P_M)$$

where $P$ is the unigram precision and $R$ is the unigram recall. The METEOR brevity penalty $P_M$ is:

$$P_M = 0.5\left(\frac{C}{M_U}\right)$$

where $C$ is the minimum number of chunks such that all unigrams in the machine translation are mapped to unigrams in the reference translation. $M_U$ is the number of unigrams that matched.

The METEOR metric incorporates a sophisticated word alignment technique that works incrementally. Each alignment stage attempts to map previously unmapped words in the SMT and reference translations. In the first phase of each stage, METEOR attempts three different types of word-to-word mappings, in the following order: exact matches, matches using stemming, and matches of synonyms. The second stage uses the resulting word mappings to evaluate word order similarity. [13]

Once a final alignment of the texts is achieved, METEOR calculates precision similar to the way the NIST metric calculates it. METEOR also calculates word-level recall between the SMT translation and the references, and combines this with precision by computing a harmonic mean that weights recall higher than precision. Lastly, METEOR penalizes shorter $n$-gram matches and rewards longer matches. [13]

TER is one of the most recent and intuitive SMT metrics developed. This metric determines the minimum number of human edits required for an SMT translation to match a reference translation in meaning and fluency. Required human edits might include inserting words, deleting words, substituting words, and changing the order or words or phrases. [14]

## 5. Experimental Results

A number of experiments were performed to evaluate various versions for our SMT systems. The experiments involved a number of steps. Processing of the corpora was accomplished, including tokenization, cleaning, factorization, conversion to lower case, splitting, and a final cleaning after splitting. Training data was processed, and the language model was developed. Tuning was performed for each experiment. Lastly, the experiments were conducted.

The baseline system testing was done using the Moses open source SMT toolkit with its Experiment Management System (EMS) [15]. The SRI Language Modeling Toolkit (SRILM) [16] with an interpolated version of the Kneser-Key discounting (interpolate –unk –kndiscount) was used for 5-gram language model training. We used the MGIZA++ tool for word and phrase alignment. KenLM [19] was used to binarize the language model, with a lexical reordering set to use the msd-bidirectional-fe model. Reordering probabilities of phrases are conditioned on lexical values of a phrase. It considers three different orientation types on source and target phrases like monotone(M), swap(S) and discontinuous(D). The bidirectional reordering model adds probabilities of possible mutual positions of source counterparts to current and following phrases. Probability distribution to a foreign phrase is determined by "f" and to the English phrase by "e" [20,21]. MGIZA++ is a multi-threaded version of the well-known GIZA++ tool [17]. The symmetrization method was set to grow-diag-final-and for word alignment processing. First two-way direction alignments obtained from GIZA++ were intersected, so only the alignment points that occurred in both alignments remained. In the second phase, additional alignment points existing in their union were added. The growing step adds potential alignment points of unaligned words and neighbors. Neighborhood can be set directly to left, right, top or bottom, as well as to diagonal (grow-diag). In the final step, alignment points between words from which at least one is unaligned are added (grow-diag-final). If the grow-diag-final-and method is used, an alignment point between two unaligned words appears. [18]

We conducted about three hundred of experiments to determine the best possible translation from Polish to English and the reverse. For experiments we used Moses SMT with Experiment Management System (EMS) [24]. Starting from baseline (BLEU: 16,02) system tests, we raised our score through extending the language model with more data and by interpolating it linearly. Firstly we used OpenSubtitles bilingual corpora for training and raised the BLEU score to 17,71. In the next step, we interpolated OpenSubtitles language model with original one instead of merging them. We determined that the linear interpolation gives better results than the log-linear one, when using our data. In the next steps, we interpolated some data and also added a Polish-English dictionary. This produces BLEU score equal to 20,41. In the PL->EN experiment number 170[th] we managed to determine better settings for the language model. We set the order from 5 to 6 and changed the discounting method from Kneser-Ney to Witten-Bell. In the training part, we changed the reordering method from msd-bidirectional-fe to msd-fe. For now, it produces the best score we were able to obtain (20,88).

As previously described, we also tried to work with stems, but the results weren't satisfying enough. Scores in fact were a bit lower – most likely because there were errors in texts. The Wroclaw NLP tools, when given a text with errors, produces even more errors and we lose some parts of the data and good alignment, which we assume is the reason for the worse score. Nevertheless, it is worth to give it a look in future research.

Because of a much bigger dictionary, the translation from EN to PL is significantly more complicated. We also lacked the data. Our baseline system score was 8,49 in BLEU. First, we tried working with stems by changing data to infinitives and reordering parts of sentences into SVO form. We then interpolated a language model containing e-books (it was prepared by us) and raised the score a little higher (9,42). Preparation of other language models and adding a bit more data resulted in achieving better scores. We also increased the n-gram order to 6, which produced BLEU score of 10,27. Next, we started to add train data (dictionary raised score to 10,40 and OpenSubtitles to 10,49) changing the alignment symmetrization method from msd-bidirectional-fe to tgttosrc (target to source) and obtained a slightly higher score. Lastly, we raised the max sentence length from 80 to 90 and achieved the highest score so far, which was 10,68 in BLEU. It must be noted that in order for the Wroclaw NLP tools to work correctly all data sets had to be previously cleaned by our tool in order to retain good alignment.

The experiments, conducted with the use of the test data from years 2010-2013, are defined in Table 1 and Table 2, respectively, for the Polish-to-English and English-to-Polish translations. They are measured by the BLEU, NIST, TER and METEOR metrics. Note that a lower value of the TER metric is better, while the other metrics are better when their values are higher. BASE stands for baseline system with no improvements, COR is a system with corrected spelling in Polish data, INF is a system using infinitive forms in Polish, SVO is a system with the subject – verb – object word order in a sentence and BEST stands for the best result we achieved.

*Table 2*: Polish-to-English translation

| System | Year | BLEU | NIST | TER | METEOR |
|--------|------|------|------|-------|--------|
| BASE | 2010 | 16.02 | 5.28 | 66.49 | 49.19 |
| COR | 2010 | 16.09 | 5.22 | 67.32 | 49.09 |
| BEST | 2010 | 20.88 | 5.70 | 64.39 | 52.74 |
| INF | 2010 | 13.22 | 4.74 | 70.26 | 46.30 |
| SVO | 2010 | 9,29 | 4,37 | 76,59 | 43,33 |
| BASE | 2011 | 18.86 | 5.75 | 62.70 | 52.72 |
| COR | 2011 | 19.18 | 5.72 | 63.14 | 52.88 |
| BEST | 2011 | 23.70 | 6.20 | 59.36 | 56.52 |
| BASE | 2012 | 15.83 | 5.26 | 66.48 | 48.60 |
| COR | 2012 | 15.86 | 5.32 | 66.22 | 49.00 |
| BEST | 2012 | 20.24 | 5.76 | 63.79 | 52.37 |
| BASE | 2013 | 16.55 | 5.37 | 65.54 | 49.99 |
| COR | 2013 | 16.98 | 5.44 | 65.40 | 50.39 |
| BEST | 2013 | 23.00 | 6.07 | 61.12 | 55.16 |
| INF | 2013 | 12.40 | 4.75 | 70.38 | 46.36 |

*Table 3*: English-to-Polish translation

| System | Year | BLEU | NIST | TER | METEOR |
|--------|------|------|------|------|--------|
| BASE | 2010 | 8.49 | 3.70 | 76.39 | 31.73 |
| COR | 2010 | 9.39 | 3.96 | 74.31 | 33.06 |
| BEST | 2010 | 10.72 | 4.18 | 72.93 | 34.69 |
| INF | 2010 | 9.11 | 4.46 | 74.28 | 37.31 |
| SVO | 2010 | 4,27 | 4,27 | 76,75 | 33,53 |
| BASE | 2011 | 10.77 | 4.14 | 71.72 | 35.17 |
| COR | 2011 | 10.74 | 4.14 | 71.70 | 35.19 |
| BEST | 2011 | 15.62 | 4.81 | 67.16 | 39.85 |
| BASE | 2012 | 8.71 | 3.70 | 78.46 | 32.50 |
| COR | 2012 | 8.72 | 3.70 | 78.57 | 32.48 |
| BEST | 2012 | 13.52 | 4.35 | 73.36 | 36.98 |
| BASE | 2013 | 9.35 | 3.69 | 78.13 | 32.52 |
| COR | 2013 | 9.35 | 3.70 | 78.10 | 32.54 |
| BEST | 2013 | 14.37 | 4.42 | 72.06 | 37.87 |
| INF | 2013 | 13.30 | 4.83 | 70.50 | 35.83 |

Official results were obtained late for this paper publication so we decided only to put our best translation system results. In translation from Polish to English, case-sensitive BLEU score was 22,60 and TER 62,56 while in case-insensitive BLEU score was equal to 23,54 and TER 61,12. In translation from English to Polish we obtained case-sensitive BLEU 14,29 and TER 73,53 while case-insensitive scores were 15,04 for BLEU and 72,05 for TER.

## 6. Discussion

Several conclusions can be drawn from the experimental results presented here. Automatic and manual cleaning of the training files has some impact, among the variations examined, on improving translation performance, together with spelling correction of the data in Polish – although it resulted in better BLEU and METEOR scores, not always in higher NIST or TER metrics. In particular, automatic cleaning and conversion of verbs to their infinitive forms improved translation performance when it comes to the English-to-Polish translation, quite the contrary to the Polish-to-English translation. This is likely due to a reduction of the Polish vocabulary size. Changing the word order to SVO is quite interesting. It didn't help at all in some cases, although one would expect it to. When it comes to experiments from PL to EN, the score was always worse, which was not anticipated. On the other hand, in the EN to PL experiments in some cases improvement could be seen. Although the BLEU score dramatically decreased and TER became slightly worse, NIST and METEOR showed better results than the baseline system. Most likely it is because of each metric has different evaluation method. The BLEU and TER scored decreased probably because phrases were mixed in the SVO conversion process. It is worth investigating (especially the PL to EN system). Maybe there was some kind of implementation error in our parser or cleaner.

In summary converting Polish verbs to infinitives reduces the Polish vocabulary, which should improve the English-to-Polish translation performance. The Polish to English translation typically outscores the English to Polish translation, even on the same data. This requires further evaluation.

## 7. Conclusion and future work

Several potential opportunities for future work are of interest. Additional experiments using extended language models are warranted to determine if this improves SMT scores. We are also interested in developing some more web crawlers in order to obtain additional data that would most likely prove useful.

Currently, neural network based language models are amongst the most successful techniques for statistical language modeling. They can be easily applied in a wide range of tasks, including automatic speech recognition and machine translation and they also provide significant improvements over classic backoff n-gram models. The 'rnnlm' toolkit can be used in order to train, evaluate and use such models. With the RNNLM toolkit we were able to reduce perplexity (Table 4) a little. We intend to explore it more in future work.

*Table 4*: RNNLM Results

| TOOLKIT | PPL |
|---------|-----|
| RNNLM | 193.31 |
| SRILM | 216.79 |
| Combination of RNNLM and SRILM | 169.55 |

The test was conducted on a default language model and a test set provided in TED 2013 data, and it looks promising. The language model vocabulary was 44052 words and the test file was 3575 words.

Polish is a language that has a complex grammar, which is why it is very hard to translate from and into languages of lower complexity such as English. Creating a factored model for SMT would most probably improve its performance. We are planning on implementing an SMT factored system based on POS tags.

An ideal SMT system should be fully automatic. To use infinitives, we will have to make this conversion automatic with usage of Wroclaw NLP tools. Lastly, it is our objective to create two SMT systems, one converting Polish words to Polish stems (and vice versa), and another converting Polish infinitives to English in order to make translations fully automatic.

The observed lower quality of the translation system based on conversion into an SVO sentence form requires further investigation.

## 8. Acknowledgements

This work is supported by the European Community from the European Social Fund within the Interkadra project UDA-POKL-04.01.01-00-014/10-00 and Eu-Bridge 7th FR EU project (grant agreement n°287658).